\documentclass{article}

\usepackage{tikz}
\usetikzlibrary{shapes.geometric, arrows.meta, positioning, fit}

  \usepackage[dblblindworkshop, final]{neurips_2025}
\workshoptitle{1st Workshop on Differentiable Systems and Scientific Machine Learning @ EurIPS 2025}

\usepackage[utf8]{inputenc} 
\usepackage[T1]{fontenc}    
\usepackage[hidelinks]{hyperref}       
\usepackage{url}            
\usepackage{booktabs}       
\usepackage{amsfonts}       
\usepackage{nicefrac}       
\usepackage{microtype}      
\usepackage{xcolor}         
\usepackage{amsmath}%
\usepackage{MnSymbol}%
\usepackage{wasysym}%
\usepackage{threeparttable}

\title{A Validated LBM Dataset and Pipeline for Surrogate Modeling of Turbulent 3D Obstructed Channel Flows}

\author{%
  Lukas Schröder \\
  Erlangen National High Performance Computing Center (NHR@FAU)\\
  \texttt{lukas.ls.schroeder@fau.de} \\
  \And
  Shubham Kavane \quad Harald Köstler \\
  Chair of Computer Science 10 (System Simulation) \\
  Friedrich-Alexander-Universität Erlangen-Nürnberg\\
  \texttt{\{shubham.kavane, harald.koestler\}@fau.de} }

\begin{document}

\maketitle

\begin{abstract}
    Evaluating neural operators for 3D turbulent flow requires validated datasets with physical benchmarks. We present a reproducible pipeline generating training data for 3D channel flows around generated geometries at Re=1,000–10,000. Our lattice Boltzmann solver with cumulant collision operators is rigorously verified against experimental measurements (Strouhal number, drag coefficients, turbulent fluctuations) with comprehensive grid convergence studies at resolution $1024\times512\times512$. Building upon an established framework, this validated pipeline enables standardized surrogate model comparison. We outline planned systematic evaluation of Fourier Neural Operator and U-Net variants on forecasting, super-resolution, and error correction tasks, using physics-informed metrics to assess turbulent energy cascade representation. Future work will compare computational efficiency between numerical solvers and neural surrogates, exploring practical application. We seek community feedback on our validation approach, planned benchmark methodology, and evaluation priorities for neural operators in turbulent flows.
\end{abstract}

\section{Introduction \& motivation}
    Computational fluid dynamics simulations of 3D turbulent flows at moderate to high Reynolds numbers remain computationally expensive. Despite ease of parallelization on modern GPUs, time-resolved predictions still require substantial computational resources. This motivates the development of machine learning surrogate models that can distill knowledge from expensive simulations into fast neural network (NN) predictions, potentially reducing computational effort while preserving essential physical accuracy. While databases such as the Johns Hopkins Turbulence Database \cite{7208756} provide high-quality simulations, they focus primarily on canonical configurations (isotropic turbulence, channel flows) rather than 3D flows around complex geometries relevant for engineering applications.

    Recent advances in neural operators, particularly Fourier Neural Operators (FNO) \cite{li2021fourier} and U-Net \cite{ronneberger2015u} architectures, have shown promise for fluid dynamics applications \cite{wang2024prediction, gonzalez2023towards}. However, rigorous comparison of state-of-the-art models remains challenging due to the lack of validated, high-quality datasets with consistent physical benchmarks. Fair model assessment requires not only comparable training data but also physically meaningful evaluation metrics, especially for turbulent flows where representation of the energy cascade and high-frequency dynamics are critical \cite{Pope_2000, oommen2025learning}. Furthermore, a fair assessment requires not only physically accurate results but also consistent measurement of computational cost from highly optimized implementations of both traditional solvers and neural models.

    This work addresses this gap by presenting a reproducible and rigorously validated pipeline for simulating flows around objects across Reynolds numbers from 1000 to 10,000. Building upon an established framework \cite{kavane2025channelflow}, which established the geometry generation and data processing infrastructure, this work introduces a rigorously validated lattice Boltzmann method (LBM) \cite{kruger2017lattice} solver configuration with cumulant collision operators and systematic experimental verification. Our LBM-based approach is verified against experimental measurements including Strouhal number, recirculation length, turbulent energy fluctuations, and drag coefficients, with grid convergence studies ensuring numerical accuracy. We provide preliminary baseline results of a Fourier Neural Operator predicting time-averaged flow fields as proof-of-concept, demonstrating the pipeline's utility. Finally, we outline our planned framework for systematic comparison of U-Net and Fourier-based Neural Operators across multiple prediction tasks. The gained insights on physics-based NNs also contribute to the research on differential solvers.

\section{Pipeline description}
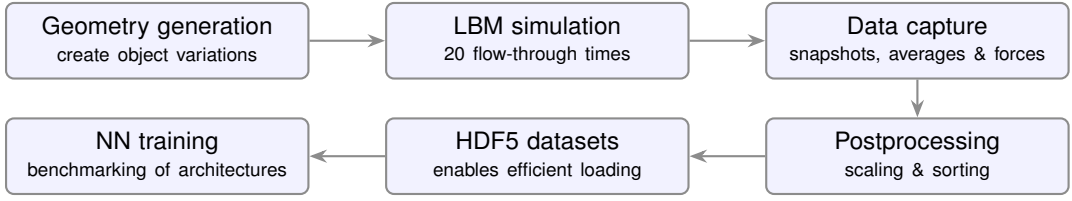
\begin{figure}[t]
        \centering
        \begin{tikzpicture} [
            node distance=0.5cm and 1cm,
            process/.style = {rectangle, minimum width=4cm, minimum height=1cm, rounded corners=3pt, text centered, draw=gray!90, thick, fill=gray!2, font=\small\sffamily, align=center, fill=blue!5},
            arrow/.style={thick,->,>=Stealth, draw=gray!90},
            ]
           \node (geom) [process, text width=3.5cm] {Geometry generation\\{\scriptsize create object variations}};
           \node (sim) [process, right=of geom, text width=3.5cm] {LBM simulation\\{\scriptsize 20 flow-through times}};
           \node (capture) [process, right=of sim, text width=3.5cm] {Data capture\\{\scriptsize snapshots, averages \& forces}};
           \node (post) [process, below=of capture, text width=3.5cm] {Postprocessing\\{\scriptsize scaling \& sorting}};
           \node (output) [process, below=of sim, text width=3.5cm] {HDF5 datasets\\{\scriptsize enables efficient loading}};
           \node (training) [process, below=of geom, text width=3.5cm] {NN training\\{\scriptsize benchmarking of architectures}};

           \draw [arrow] (geom) -- (sim);
           \draw [arrow] (sim) -- (capture);
           \draw [arrow] (capture) -- (post);
           \draw [arrow] (post) -- (output);
           \draw [arrow] (output) -- (training);
        \end{tikzpicture}
        \caption{Pipeline workflow from geometry generation for NN training \cite{kavane2025channelflow}}
        \label{fig:pipeline}
    \end{figure}

    The pipeline in Figure~\ref{fig:pipeline} provides code that handles data generation, simulation, and curation for ready-to-use results in NN training. Starting with a module that generates derivative objects from a set of primitives $[\text{Cylinder}, \text{Rectangle}, \text{Sphere}, \text{Torus}, \text{Wedge}]$, which can also be replaced by user-provided objects. The flow around objects can then be simulated in the defined Reynolds number range $[1000, 10000]$ with varying angles of attack using a LBM simulation (detailed in Section~\ref{sec:solver}). The simulation captures the average flow after 15 flow-through times, recording the mean velocity and pressure of the flow, together with 20 consecutive snapshots separated by 25 $\delta t$ in lattice units. Additionally, the pipeline records the accumulated forces on the object in x and y directions and 2D snapshots of the center x-y plane for monitoring the development of the entire simulation. Finally, the compressed files per simulation can be unpacked, scaled, and repacked into HDF5 files with specified resolutions in the range of $[(512 \times 256 \times 256), (64 \times 32 \times 32)]$, along with the additional flag field, which captures the spatial occupancy mask or a signed distance field. 

\section{Dataset creation and validation}
    \subsection{Generated dataset}
        The preliminary dataset comprises flows around 42 procedurally generated objects within a computational domain of dimensions $2 \times 1 \times 1$ (in units of channel height $L$). Each object is positioned at $x/2L = 0.3$, centered in the y-z plane and spans $1/6L$ (86 cells). Reynolds numbers $Re_c$ from 1000 to 10,000, in the transitional-to-turbulent regime, are achieved by adjusting the lattice viscosity while maintaining constant inlet velocity, with Re defined based on channel height. The inflow is implemented using a prescribed velocity boundary condition, the outflow employs extrapolation, and the lateral walls enforce no-slip conditions. The domain is resolved at $1024 \times 512 \times 512$ to ensure quality, as mesh refinement in LBM reduces accuracy due to interpolation artifacts. Data is stored at half resolution; simulations require 3h on 4 Nvidia H100 GPUs.

    \subsection{waLBerla framework and solver setup}
    \label{sec:solver}
        Simulations employ the waLBerla framework (licensed under \href{https://i10git.cs.fau.de/walberla/walberla/-/blob/master/COPYING.txt}{GPLv3}), which provides highly optimized code-generated LBM implementations with efficient parallelization capabilities \cite{bauer2021walberla}. This framework's performance characteristics allow for a fair computational effort comparison between it and surrogates later.

        The cumulant collision operator with a D3Q27 stencil provides improved numerical stability and accuracy compared to moment-based multiple relaxation time models \cite{geier2015cumulant}. A fourth-order advection-diffusion correction scheme with limiting enhances numerical precision and stability for turbulent flows \cite{geier2017parametrization}. This approach accurately captures the energy cascade and maintains well-resolved energy spectra essential for turbulence representation \cite{Pope_2000}. The method's inherent controlled dissipation effectively models subgrid-scale effects, eliminating the need for explicit subgrid models (SGS) such as Smagorinsky closures. Testing confirmed that adding SGS modeling reduced physical accuracy. Notably, the correction introduces instabilities in laminar flows with elevated kinematic viscosity, so the correction is disabled for $Re_c<3000$.
        Fluid-solid interaction at object boundaries employs a quadratic bounce-back scheme, which provides second-order spatial accuracy at the cost of computing wall distances \cite{geier2015cumulant}. This approach ensures physically accurate boundary conditions critical for turbulent flow development.

    \subsection{Validation}
    \label{sec:validation}
        The primary objective of this dataset is providing physically accurate training data that enables rigorous assessment of NN surrogates' ability to capture turbulent flow physics. Comprehensive validation was therefore conducted across multiple physical parameters representing the dynamics and the average properties of the flow.

        Turbulent regime validation uses the experimental measurements of Choi and Park \cite{choi2018flow} for flow around a sphere at $Re_s$ 1000, relative to sphere diameter $D$, corresponding to $Re_c$ 6000 relative to our channel height. This reference provides validation of Strouhal number, recirculation length, and turbulent velocity fluctuation statistics.

        To verify dataset validity extends to laminar conditions, drag coefficient validation was performed at the Reynolds number extremes: $Re_c$ 600 and $Re_c$ 12000 even outside our simulated range. Additionally, a grid convergence study \cite{roache1997quantification} on drag coefficients confirms numerical accuracy at our chosen resolution. Validation results and quantitative comparisons are presented in the following, with all the plots and referenced data in the appendix \ref{sec:val_plots}.

        \textbf{Drag Coefficient \& Grid Convergence:} The drag coefficient is an expressive and simple to conduct measurement, because it summarizes the pressure distribution, viscous effects and separation and wake dynamics in one parameter while being sensitive to numerical errors as well. The coefficients show a convergent behavior in Table~\ref{tab:drag_coefficient_data}, matching experimental values within $\pm 6\%$ of Roos and Willmarth \cite{roos1971some} with our final spacing of 0.0019 ($L \sim 512$ cells). The error growth with $Re_s$ above 1000, which demands more attention in future research.

        \begin{table}[htbp]
            \caption{Drag coefficients at different Reynolds numbers and grid resolutions}
            \label{tab:drag_coefficient_data}
            \centering
            \begin{tabular}{llcccccl}
                \toprule
                \multicolumn{2}{c}{Reynolds number} & \multicolumn{5}{c}{drag coefficient at resolution} & \\
                \cmidrule(lr){1-2}\cmidrule(lr){3-7}
                $\mathrm{Re}_s$ & $\mathrm{Re}_c$ & 128 & 256 & 384 & 512 & 1024 & reference \cite{roos1971some} \\
                \midrule
                100  & 600   & 1.122 & 1.108 & 1.106 & 1.106 & ---   & 1.08  \\
                500  & 3000  & 0.573 & 0.562 & 0.563 & 0.562 & ---   & 0.547 \\
                1000 & 6000  & 0.479 & 0.460 & 0.466 & 0.467 & 0.472 & 0.472 \\
                2000 & 12000 & 0.434 & 0.407 & 0.407 & 0.404 & ---   & 0.430 \\
                \bottomrule
            \end{tabular}
        \end{table}

        \textbf{Average Field with recirculation length:} The normalized mean streamwise velocity field in Figure~\ref{fig:u_mean_path} exhibits a recirculation bubble of $l_r/D = 1.8$ at $Re_s$ 1000, consistent with the measured range of $1.76 \pm 0.07$ of Choi and Park \cite{choi2018flow}. The velocity profiles in Figure~\ref{fig:avg_velocity_comparison} and \ref{fig:avg_vertical_velocity_comparison} closely match the measured mean flow distribution, confirming accurate capture of the mean flow characteristics.

        \textbf{Strouhal number:} Spectral analysis of vertical velocity fluctuations in the wake reveals a dominant frequency corresponding to $St \approx 0.18$ at $Re_s$ 1000 in Figure~\ref{fig:psd}, agreeing with literature values and confirming accurate capture of vortex shedding dynamics.

        \textbf{Streamwise turbulent intensity:} This serves as an indicator of the scale of the fluctuations relative to the distance and thus is essential in capturing the mechanics of turbulent energy dissipation. Figure~\ref{fig:trub_intensity} shows two peaks 1.5 and 2.4 $x/D$ which is identical to the experiment of Choi and Park \cite{choi2018flow} with the scale in between their measurements and their referenced sources.

        \textbf{Turbulent structures:} The instantaneous snapshot in Figure~\ref{fig:snapshot} reproduce characteristic turbulent features including hairpin vortices with identifiable heads and trailing legs, validating the solver's capability to resolve fine-scale turbulent coherent structures. 

        The validated solver configuration is applied unchanged to diverse geometries, consistent with established CFD surrogate modeling practice, where canonical benchmarking validates numerical frameworks for application across geometrically diverse cases.

\section{Baseline results}
    We present preliminary baseline results in Table~\ref{tab:first_results} for predicting the mean streamwise velocity field from a binary geometry mask using factorized FNO \cite{tran2021factorized}, original FNO \cite{li2021fourier}, a combination of both, and a U-Net \cite{ronneberger2015u} model as a proof-of-concept. These baselines establish reference levels of predictive accuracy and computational efficiency, serving as a foundation for evaluating and comparing future improvements in surrogate modeling. The U-Net outperforms the other base models and thus can be set as the most promising candidate for further research. The settings are listed under appendix \ref{sec:baseline}. 

    \begin{table}[htbp]
        \centering
        \caption{Preliminary baseline results on validation data (details in appendix \ref{sec:baseline})}
        \label{tab:first_results}
        \begin{tabular}{lllllll}
            \toprule
            model & layers & MSE & max MAE & NRMSE & MAPE & sample/s \\
            \midrule
            FNO & 4 & $6.01 \cdot 10^{-4}$ & 0.0338 & 0.0359 & 5.71 & 20.77\\
            Hybrid F/FNO & 4 & $6.69 \cdot 10^{-4}$ & 0.0345 & 0.0379 & 5.80 & 16.36\\
            FFNO & 4 & $11.13 \cdot 10^{-4}$ & 0.0464 & 0.0495 & 7.50 & 12.30\\
            U-Net & 5 & $1.22 \cdot 10^{-4}$ & 0.0158 & 0.0159 & 2.17 & 32.87 \\
            \bottomrule
        \end{tabular}
        \end{table}

\section{Planned model comparison}
    Having established a validated dataset, we now focus on systematic model comparison. The experimental setup covers fundamental surrogate modeling tasks: (1) predicting mean flow fields from geometry and Reynolds number, and (2) next-timestep forecasting from current state. When architectures support it, we will extract drag coefficients ($c_d$), perform super-resolution from low-resolution snapshots, and evaluate temporality through multi-timestep input/output configurations.

    From a methodological perspective, we focus on advancements of baseline U-Nets and FNOs. A critical challenge in turbulent flow prediction is spectral bias—the tendency to oversmooth fine-scale structures \cite{rahaman2019spectral}. Following Oommen et al. \cite{oommen2025learning}, we plan to evaluate expanded loss functions combining adversarial and spectral components alongside standard $L_2$ losses. We will apply these techniques to Fourier-based architectures, U-Net variants, and hybrid combinatory models \cite{wen2022u, gonzalez2023towards}. Additionally, exploring attention mechanisms integrated as channel/spatial gates in Time-Conditioned U-Nets (\cite{ovadia2025real}) and as linear attention coupling in FNOs (\cite{peng2023linear}).
    Finally, rigorous comparison of computational costs between solvers and surrogates will assess practical applicability in 3D turbulent flows.

\section{Limitations}
    One of the limitations of our pipeline is the high computational and storage demand of high-resolution turbulent data generation, limiting dataset and resolution scalability. The solver is limited for the transitional-to-turbulent regime (Re 1,000–10,000). The baseline results lack variance estimation or comprehensive hyperparameter tuning, are predicting a significantly lower resolution with reduced turbulent scales influence and serve only as proof-of-concept demonstrations.

\section{Conclusion}
    This work-in-progress presents a validated pipeline for generating physically accurate 3D turbulent flow datasets around complex geometries. We welcome feedback from the community on our planned systematic comparison of U-Net and Fourier Neural Operator variants, particularly regarding methodological priorities and evaluation metrics for flow prediction surrogates. Ultimately, this validated pipeline aims to advance physics-based NN, also as building block of hybrid differentiable numerical solvers.

\begin{ack}
    The authors gratefully acknowledge the HPC resources provided by the Erlangen National High Performance Computing Center (NHR@FAU) of the Friedrich-Alexander-Universität Erlangen-Nürnberg (FAU). NHR funding is provided by the German Federal Ministry of Education and Research and the state governments participating on the basis of the resolutions of the GWK for the national high-performance computing at universities (www.nhr-verein.de/unsere-partner) by federal and Bavarian state authorities. NHR@FAU hardware is partially funded by the German Research Foundation (DFG) – 440719683
\end{ack}


\bibliographystyle{unsrt}
\bibliography{refs}

@article{7208756,
  author={Kanov, Kalin and Burns, Randal and Lalescu, Cristian and Eyink, Gregory},
  journal={Computing in Science \& Engineering}, 
  title={The Johns Hopkins Turbulence Databases: An Open Simulation Laboratory for Turbulence Research}, 
  year={2015},
  volume={17},
  number={5},
  pages={10-17},
  doi={10.1109/MCSE.2015.103}}

@article{bauer2021walberla,
  title={waLBerla: A block-structured high-performance framework for multiphysics simulations},
  author={Bauer, Martin and Eibl, Sebastian and Godenschwager, Christian and Kohl, Nils and Kuron, Michael and Rettinger, Christoph and Schornbaum, Florian and Schwarzmeier, Christoph and Th{\"o}nnes, Dominik and K{\"o}stler, Harald and others},
  journal={Computers \& Mathematics with Applications},
  volume={81},
  pages={478--501},
  year={2021},
  publisher={Elsevier}
}

@article{geier2015cumulant,
  title={The cumulant lattice Boltzmann equation in three dimensions: Theory and validation},
  author={Geier, Martin and Sch{\"o}nherr, Martin and Pasquali, Andrea and Krafczyk, Manfred},
  journal={Computers \& Mathematics with Applications},
  volume={70},
  number={4},
  pages={507--547},
  year={2015},
  publisher={Elsevier}
}

@article{geier2017parametrization,
  title={Parametrization of the cumulant lattice Boltzmann method for fourth order accurate diffusion part I: Derivation and validation},
  author={Geier, Martin and Pasquali, Andrea and Sch{\"o}nherr, Martin},
  journal={Journal of Computational Physics},
  volume={348},
  pages={862--888},
  year={2017},
  publisher={Elsevier}
}

@article{choi2018flow,
  title={Flow around in-line sphere array at moderate Reynolds number},
  author={Choi, Daehyun and Park, Hyungmin},
  journal={Physics of Fluids},
  volume={30},
  number={9},
  year={2018},
  publisher={AIP Publishing}
}

@article{kavane2025channelflow,
  title={ChannelFlow-Tools: A Standardized Dataset Creation Pipeline for 3D Obstructed Channel Flows},
  author={Kavane, Shubham and Kulkarni, Kajol and Koestler, Harald},
  journal={arXiv preprint arXiv:2509.15236},
  year={2025}
}

@article{oommen2025learning,
  title={Learning Turbulent Flows with Generative Models: Super-resolution, Forecasting, and Sparse Flow Reconstruction},
  author={Oommen, Vivek and Khodakarami, Siavash and Bora, Aniruddha and Wang, Zhicheng and Karniadakis, George Em},
  journal={arXiv preprint arXiv:2509.08752},
  year={2025}
}

@article{roos1971some,
  title={Some experimental results on sphere and disk drag},
  author={Roos, Frederick W and Willmarth, William W},
  journal={AIAA journal},
  volume={9},
  number={2},
  pages={285--291},
  year={1971}
}

@article{wen2022u,
  title={U-FNO—An enhanced Fourier neural operator-based deep-learning model for multiphase flow},
  author={Wen, Gege and Li, Zongyi and Azizzadenesheli, Kamyar and Anandkumar, Anima and Benson, Sally M},
  journal={Advances in Water Resources},
  volume={163},
  pages={104180},
  year={2022},
  publisher={Elsevier}
}

@article{peng2023linear,
  title={Linear attention coupled Fourier neural operator for simulation of three-dimensional turbulence},
  author={Peng, Wenhui and Yuan, Zelong and Li, Zhijie and Wang, Jianchun},
  journal={Physics of Fluids},
  volume={35},
  number={1},
  year={2023},
  publisher={AIP Publishing}
}

@article{ovadia2025real,
  title={Real-time inference and extrapolation with Time-Conditioned UNet: Applications in hypersonic flows, incompressible flows, and global temperature forecasting},
  author={Ovadia, Oded and Oommen, Vivek and Kahana, Adar and Peyvan, Ahmad and Turkel, Eli and Karniadakis, George Em},
  journal={Computer Methods in Applied Mechanics and Engineering},
  volume={441},
  pages={117982},
  year={2025},
  publisher={Elsevier}
}

@article{tran2021factorized,
  title={Factorized fourier neural operators},
  author={Tran, Alasdair and Mathews, Alexander and Xie, Lexing and Ong, Cheng Soon},
  journal={arXiv preprint arXiv:2111.13802},
  year={2021}
}

@article{li2021fourier,
  title={Fourier neural operator for parametric partial differential equations},
  author={Li, Zongyi and Kovachki, Nikola and Azizzadenesheli, Kamyar and Liu, Burigede and Bhattacharya, Kaushik and Stuart, Andrew and Anandkumar, Anima},
  journal={arXiv preprint arXiv:2010.08895},
  year={2021},
  eprint={2010.08895},
  archivePrefix={arXiv},
  primaryClass={cs.LG},
  url={https://arxiv.org/abs/2010.08895}, 
}

@inproceedings{ronneberger2015u,
  title={U-net: Convolutional networks for biomedical image segmentation},
  author={Ronneberger, Olaf and Fischer, Philipp and Brox, Thomas},
  booktitle={International Conference on Medical image computing and computer-assisted intervention},
  pages={234--241},
  year={2015},
  organization={Springer}
}

@book{Pope_2000,
    place={Cambridge},
    title={Turbulent Flows},
    publisher={Cambridge University Press}, 
    author={Pope, Stephen B.},
    year={2000}
}

@book{kruger2017lattice,
  title={The lattice Boltzmann method},
  author={Kr{\"u}ger, Timm and Kusumaatmaja, Halim and Kuzmin, Alexandr and Shardt, Orest and Silva, Goncalo and Viggen, Erlend Magnus},
  volume={10},
  number={978-3},
  year={2017},
  publisher={Springer}
}

@inproceedings{rahaman2019spectral,
  title={On the spectral bias of neural networks},
  author={Rahaman, Nasim and Baratin, Aristide and Arpit, Devansh and Draxler, Felix and Lin, Min and Hamprecht, Fred and Bengio, Yoshua and Courville, Aaron},
  booktitle={International conference on machine learning},
  pages={5301--5310},
  year={2019},
  organization={PMLR}
}

@article{gonzalez2023towards,
  title={Towards Long-Term predictions of Turbulence using Neural Operators},
  author={Gonzalez, Fernando and Demoulin, Fran{\c{c}}ois-Xavier and Bernard, Simon},
  journal={arXiv preprint arXiv:2307.13517},
  year={2023}
}

@article{wang2024prediction,
  title={Prediction of turbulent channel flow using Fourier neural operator-based machine-learning strategy},
  author={Wang, Yunpeng and Li, Zhijie and Yuan, Zelong and Peng, Wenhui and Liu, Tianyuan and Wang, Jianchun},
  journal={Physical Review Fluids},
  volume={9},
  number={8},
  pages={084604},
  year={2024},
  publisher={APS}
}

@article{roache1997quantification,
  title={Quantification of uncertainty in computational fluid dynamics},
  author={Roache, Patrick J},
  journal={Annual review of fluid Mechanics},
  volume={29},
  number={1},
  pages={123--160},
  year={1997},
  publisher={Annual Reviews 4139 El Camino Way, PO Box 10139, Palo Alto, CA 94303-0139, USA}
}

@article{sitzmann2020implicit,
  title={Implicit neural representations with periodic activation functions},
  author={Sitzmann, Vincent and Martel, Julien and Bergman, Alexander and Lindell, David and Wetzstein, Gordon},
  journal={Advances in neural information processing systems},
  volume={33},
  pages={7462--7473},
  year={2020}
}


\newpage
\appendix
\section{Technical Appendices and Supplementary Material}
\subsection{Validation plots and references}
\label{sec:val_plots}

    \begin{figure}[h!]
        \centering
        \includegraphics[width=\textwidth]{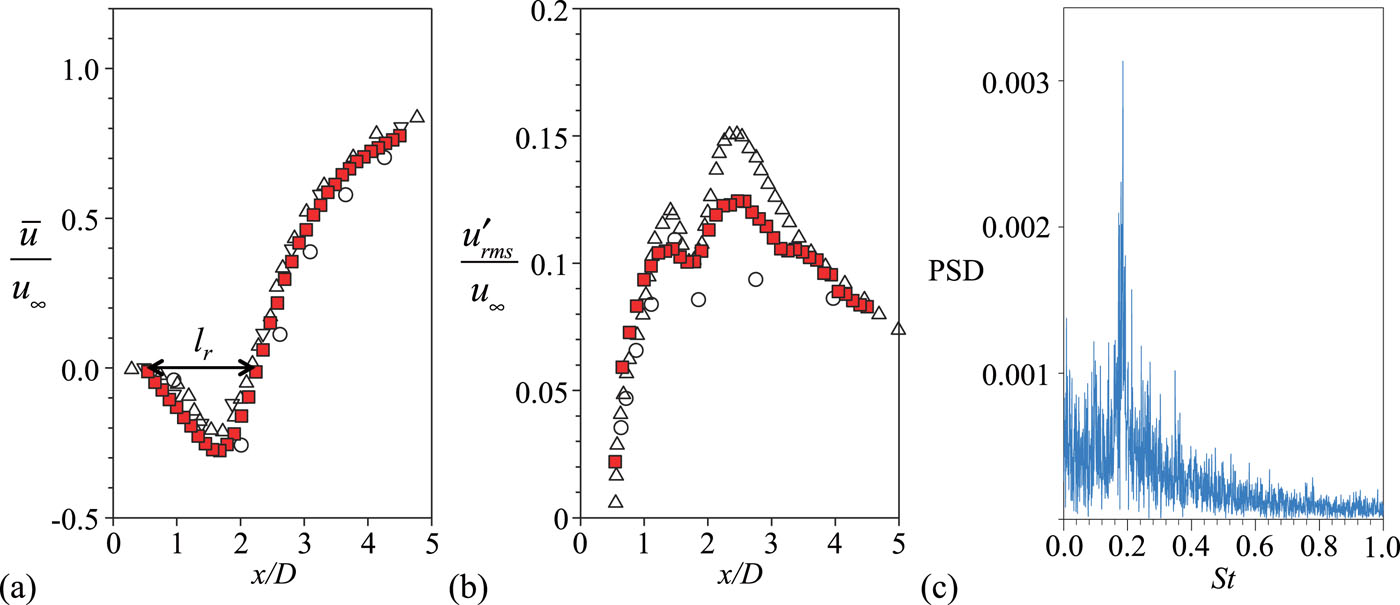}
        \caption{Reference measurements from Choi et al. of "wake characteristics behind a single sphere at Re = 1000: (a) normalized mean streamwise velocity; (b) streamwise turbulence intensity along the centerline; (c) power spectral density (PSD) of the vertical velocity" (reference measurements: $\blacksquare$, rest: literature comparisons from Choi and Park \cite{choi2018flow})}
        \label{fig:reference_plots}
    \end{figure}

    \begin{figure}[h!]
        \centering
        \includegraphics[width=0.65\textwidth]{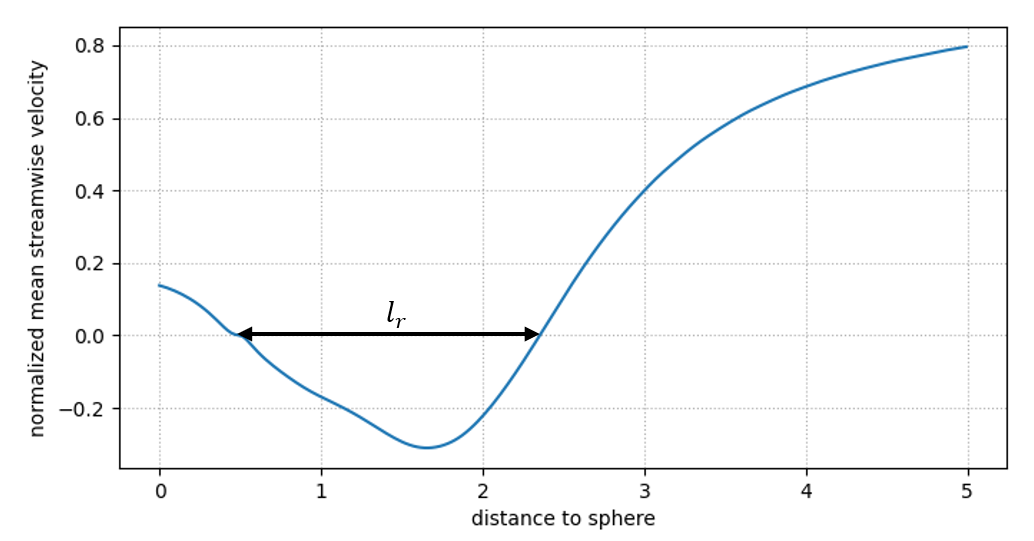}
        \caption{Normalized mean streamwise velocity $u/U_\infty$ showing recirculation length $l_r/D = 1.8$ at $Re_s$ = 1000}

        \label{fig:u_mean_path}
    \end{figure}

    \begin{figure}[h!]
        \centering
        \includegraphics[width=0.7\textwidth]{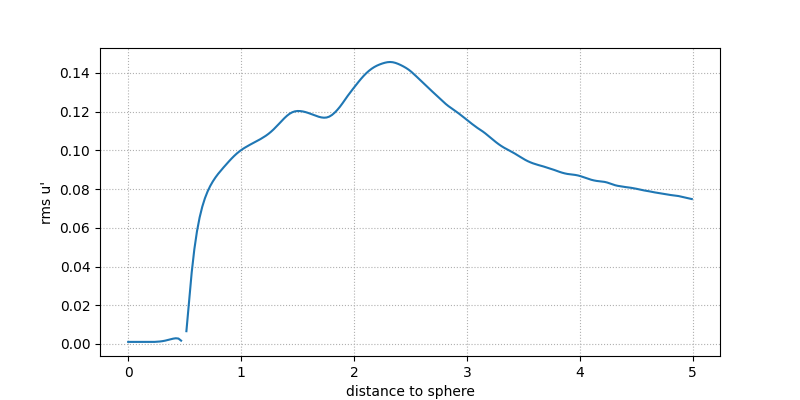}
        \caption{RMS of streamwise velocity fluctuations $u'_{rms}/U_\infty$ along centerline at $Re_s$ = 1000. Two peaks at $x/D = 1.5$ and $2.4$ match reference \cite{choi2018flow}.}
        \label{fig:trub_intensity}
    \end{figure}

    \begin{figure}[h!]
        \centering
        \includegraphics[width=0.65\textwidth]{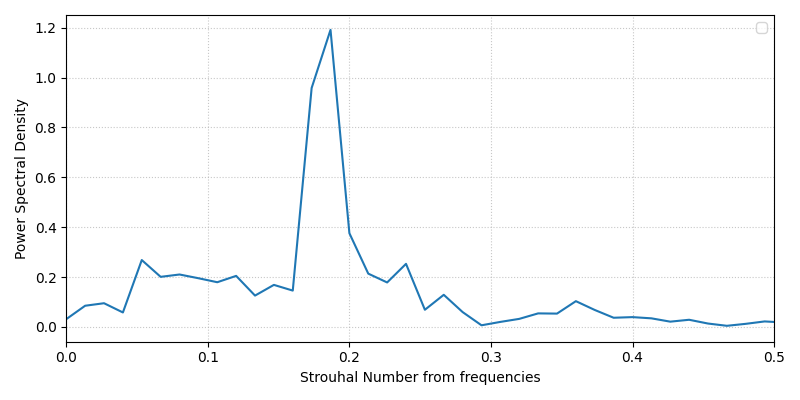}
        \caption{Power spectral density of vertical velocity showing dominant frequency at St $\approx$ 0.18, consistent with experimental vortex shedding frequency for $Re_s$ = 1000 \cite{choi2018flow}}
        \label{fig:psd}
    \end{figure}

    \begin{figure}[h!]
        \centering
        \hspace{0.7cm}\includegraphics[width=0.85\textwidth]{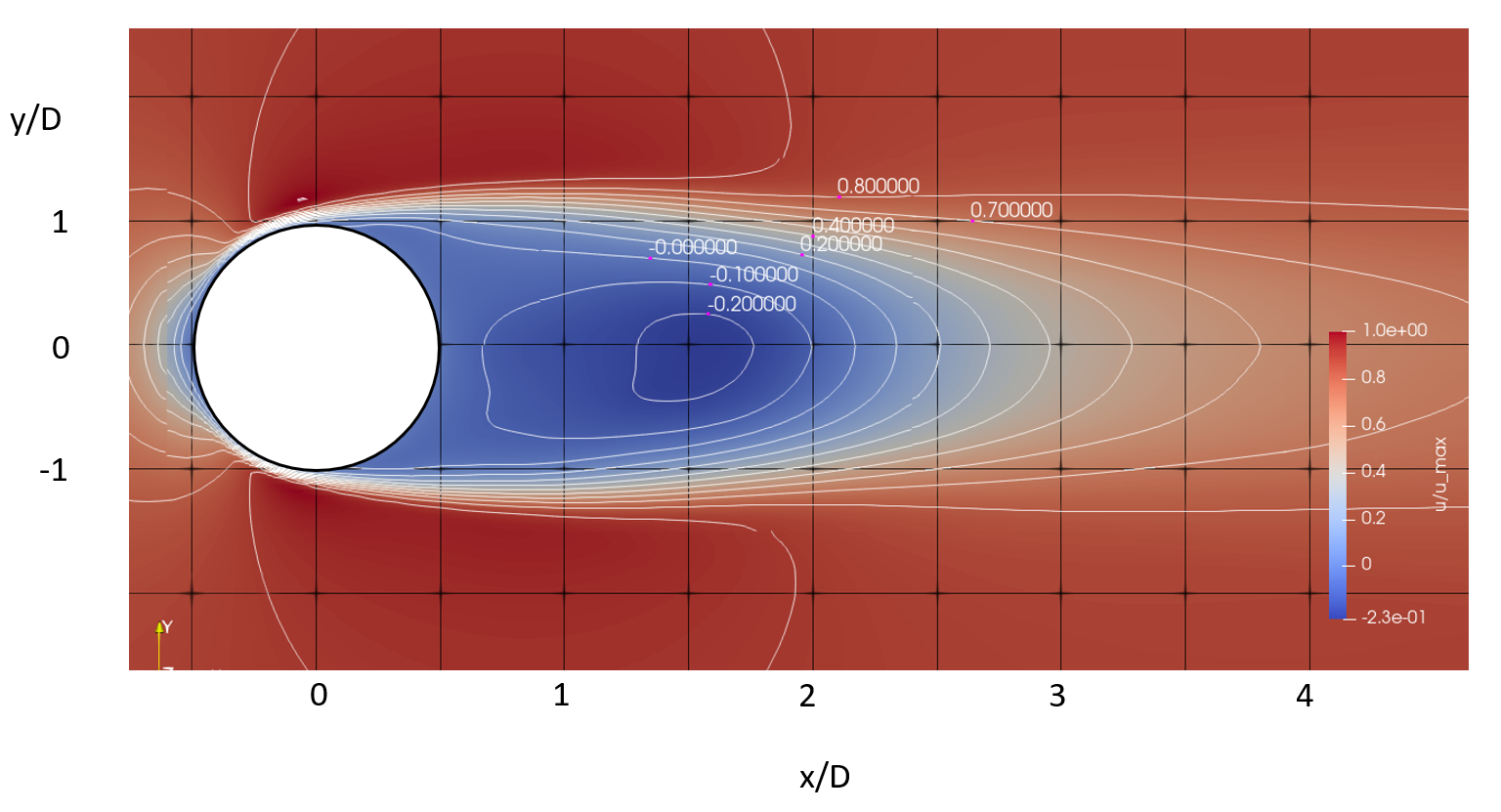}
        \vspace{0.1cm}
        \textbf{(a)}
        \vspace{0.4cm}
    
        \includegraphics[width=0.9\textwidth]{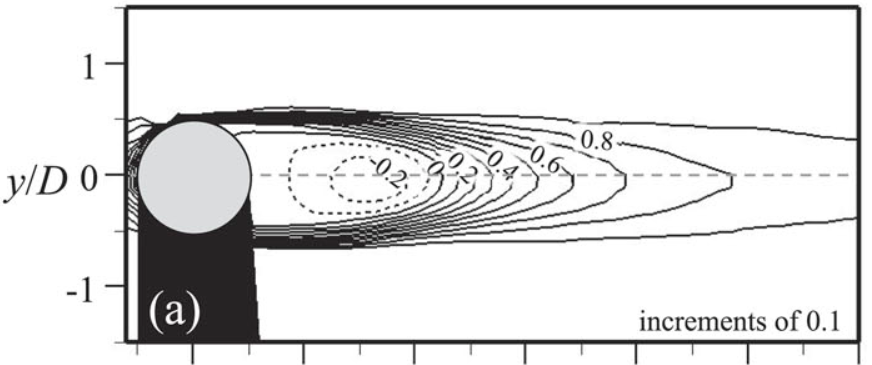}
        \vspace{0.1cm}
        \textbf{(b)}
        \caption{Mean streamwise velocity field $\langle u \rangle / U_\infty$ at $Re_s$ = 1000 validation. Contour lines show velocity magnitude. (a) LBM simulation with cumulant collision operator. (b) PIV experimental data from Choi \& Park \cite{choi2018flow}.}
        \label{fig:avg_velocity_comparison}
    \end{figure}

    \begin{figure}[h!]
        \centering
        \hspace{0.5cm}\includegraphics[width=0.88\textwidth]{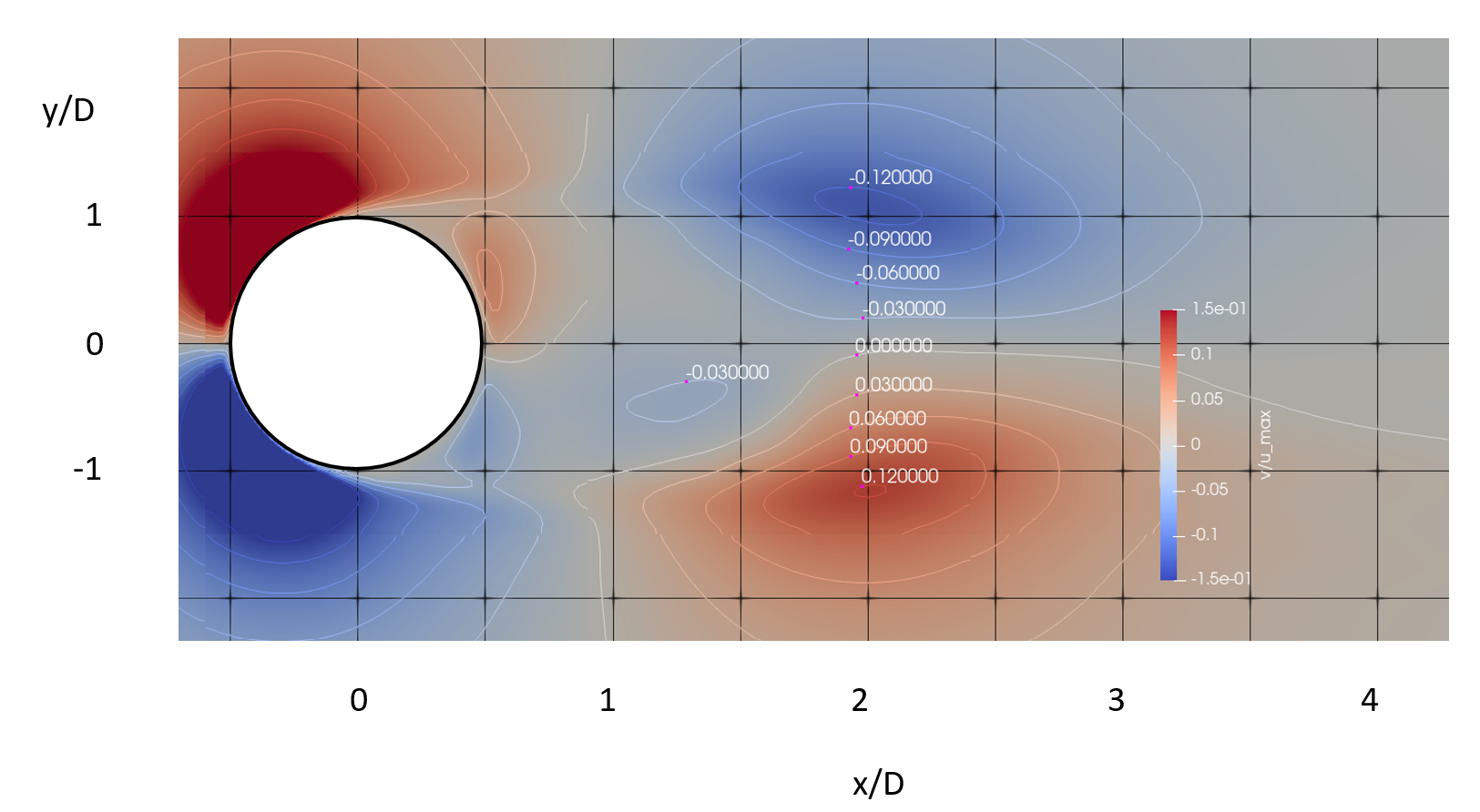}
        \vspace{0.1cm}
        \textbf{(a)}
        \vspace{0.4cm}
    
        \includegraphics[width=0.9\textwidth]{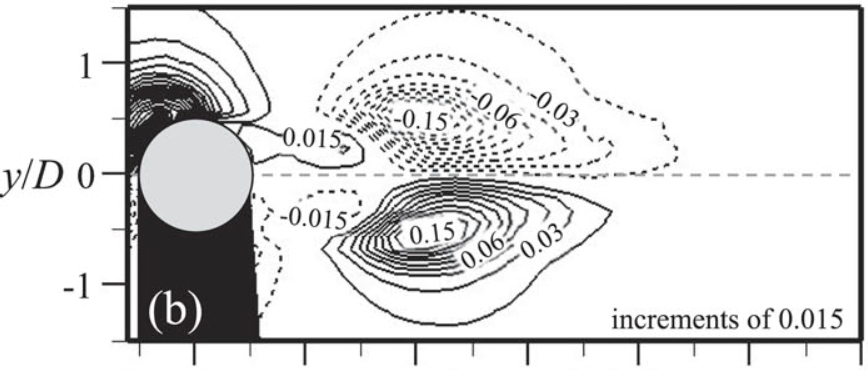}
        \vspace{0.1cm}
        \textbf{(b)}
        \caption{Mean vertical velocity field $\langle v \rangle / U_\infty$ at $Re_s$ = 1000 validation. Contour lines show velocity magnitude. (a) LBM simulation with cumulant collision operator. (b) PIV experimental data from Choi \& Park \cite{choi2018flow}.}
        \label{fig:avg_vertical_velocity_comparison}
    \end{figure}

    \begin{figure}[h!]
        \centering
        \includegraphics[width=0.9\textwidth]{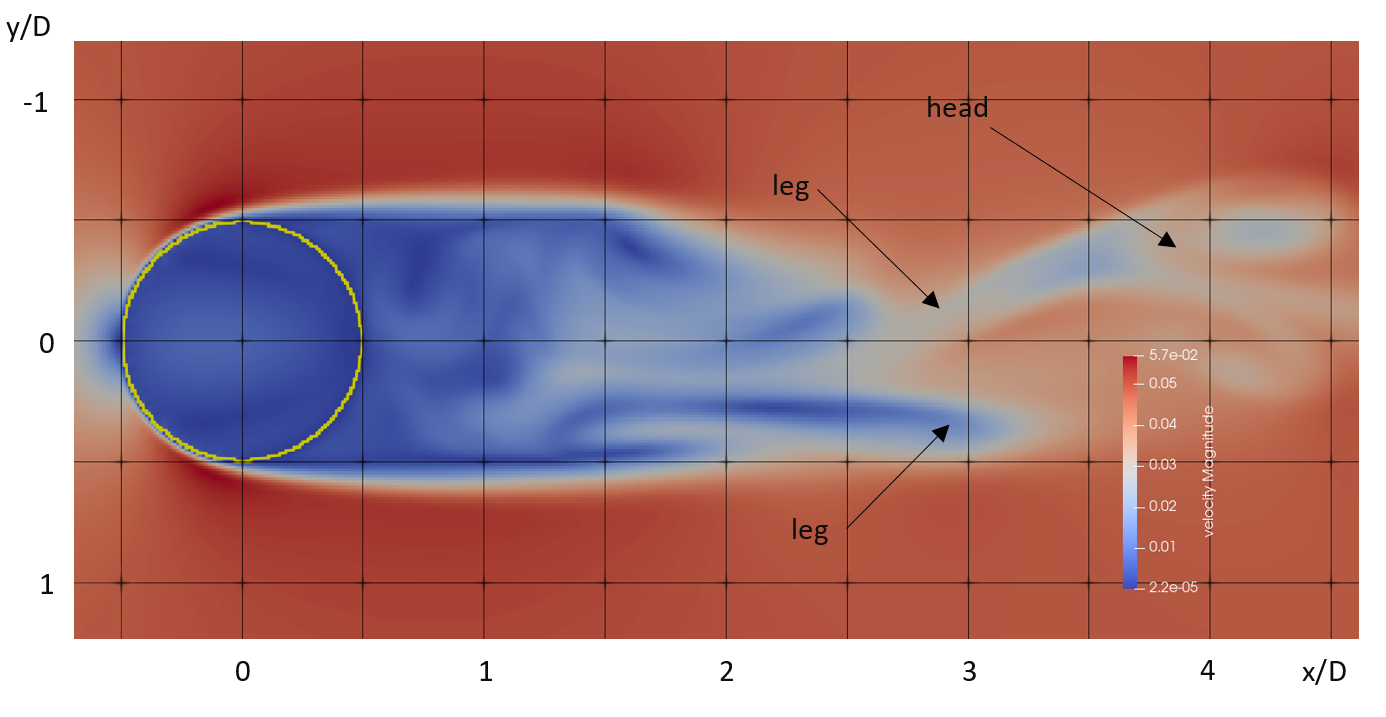}
        \caption{Instantaneous turbulent structures visualized showing characteristic hairpin vortices with identifiable heads and trailing legs at $Re_s$ = 1000.}
        \label{fig:snapshot}
    \end{figure}

    \newpage

    \subsection{Dataset and training details of the baseline results}
    \label{sec:baseline}
        The preliminary dataset contains flow around 42 geometries each simulated in the Reynolds number range of [2000, .., 6000, 8000, 9000, 10000]. 3 rotations of $90^{\circ}$ around the x-axis per geometry have been added to augment the existing data. The resolution was scaled down to $128\times64\times64$ with min-max-scaling applied for each field. The input consists of a binary field and the Re number, the output of the 3-component velocity and the density field. A split 0.8/0.1/0.1 into train, test and validation set was applied. Mean square error with masking, focusing on the important and complex inner region $([(0.2\times0.25\times0.25), (1.8\times0.75\times0.75)]$ of $(2\times1\times1))$, was chosen as the loss function. The training endured 50 epochs with the Adam optimizer and cosine learning rate scheduling and the metric of samples/s of the training on Nvidia A100 GPUs was reported and the training time for all models was below 2h.

        The FNO models represent $32\times16\times16$ modes in the frequency space and a depth of 32 channels in the co-domain. The lifting layer was set to a simple feed forward network to expand the input to the 32 channels. The projection layers is chosen as the SIREN from Sitzman et al. \cite{sitzmann2020implicit}. The FFNO follows the structure of the Tran et al. \cite{tran2021factorized} and the F/FNO hybrid maintains the FFNO structure, while replacing the split Fourier transformation with the original 3D, without any loss of practical computational time.

        The U-Net follows a standard 3D encoder-decoder architecture with skip connections. The encoder comprises four downsampling stages with channel dimensions [32, 64, 128, 256], each employing two convolutional blocks with batch normalization and ReLU activation, followed by max-pooling. The bottleneck operates at 512 channels, while the decoder mirrors the encoder with upsampling via trilinear interpolation and concatenation with skip connections. A final $1\times1\times1$ convolution projects to the output channels with sigmoid activation.

        Models are evaluated in Table~2 by the mean square error (MSE), the sample of highest mean absolute error (MAE), normalized root MSE (NRMSE) and mean absolute percentage error (MAPE).

        \begin{figure}[h!]
            \centering
            \includegraphics[width=\textwidth]{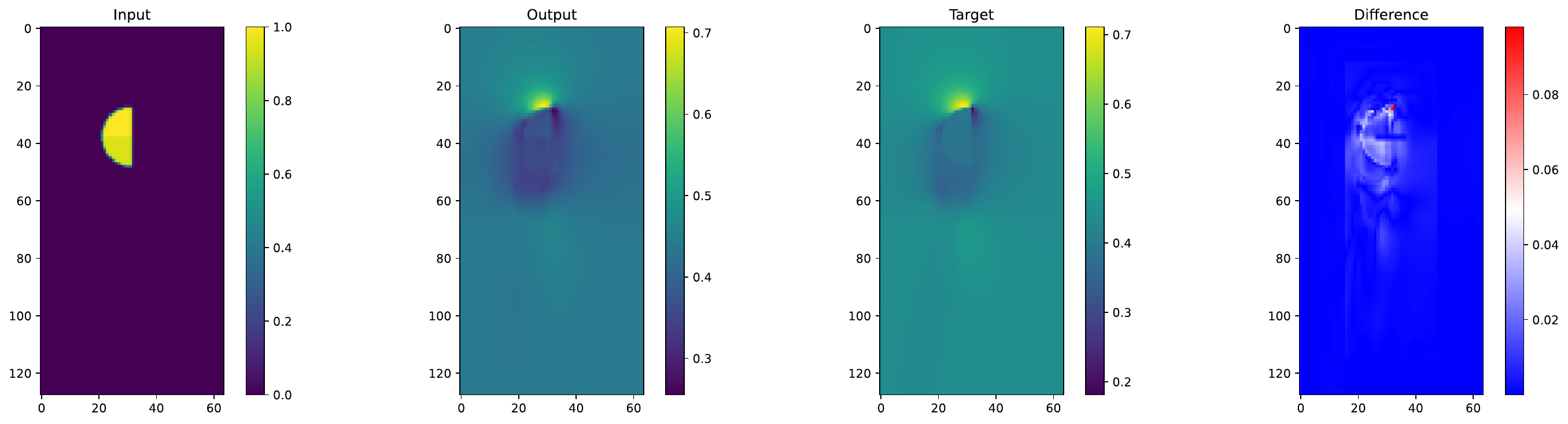}
            \caption{Slice plot of the middle plane of x-direction velocity of the baseline UNet model from the validation set}
            \label{fig:UNet}
        \end{figure}

\end{document}